\documentclass[sn-mathphys,Numbered]{sn-jnl}% Math and Physical Sciences Reference Style
\usepackage{graphicx}%
\usepackage{multirow}%
\usepackage{amsmath,amssymb,amsfonts}%
\usepackage{amsthm}%
\usepackage{array}
\usepackage{mathrsfs}%
\usepackage[title]{appendix}%
\usepackage{xcolor}%
\usepackage{textcomp}%
\usepackage{manyfoot}%
\usepackage{booktabs}%
\usepackage{algorithm}%
\usepackage{algorithmicx}%
\usepackage{algpseudocode}%
\usepackage{listings}%
\usepackage{adjustbox}
\theoremstyle{thmstyleone}%
\theoremstyle{thmstyletwo}%

\theoremstyle{thmstylethree}%

\begin{document}

\title{Enhancing the vision-language foundation model with key semantic knowledge-emphasized report refinement}

\author[1,2,3]{\fnm{Weijian} \sur{Huang}}
\author[1]{\fnm{Cheng} \sur{Li}}
\author[1,2,3]{\fnm{Hao} \sur{Yang}}
\author[1,2,3]{\fnm{Jiarun} \sur{Liu}}
\author[2]{\fnm{Yong} \sur{Liang}}
\author[1]{\fnm{Hairong} \sur{Zheng}}
\author*[1,2]{\fnm{Shanshan} \sur{Wang}}\email{ss.wang@siat.ac.cn}

\affil[1]{\orgname{Paul C. Lauterbur Research Center for Biomedical Imaging, Shenzhen Institute of Advanced Technology}, \orgaddress{\city{Shenzhen}, \country{China}}}
\affil[2]{\orgname{Pengcheng Laboratory}, \orgaddress{\city{Shenzhen}, \country{China}}}
\affil[3]{\orgname{University of Chinese Academy of Sciences}, \orgaddress{\city{Beijing}, \country{China}} \\ \\ Weijian Huang and Cheng Li contributed equally to this work.}

\abstract{Recently, vision-language representation learning has made remarkable advancements in building up medical foundation models, holding immense potential for transforming the landscape of clinical research and medical care. The underlying hypothesis is that the rich knowledge embedded in radiology reports can effectively assist and guide the learning process, reducing the need for additional labels. However, these reports tend to be complex and sometimes even consist of redundant descriptions that make the representation learning too challenging to capture the key semantic information. This paper develops a novel iterative vision-language representation learning framework by proposing a key semantic knowledge-emphasized report refinement method. Particularly, raw radiology reports are refined to highlight the key information according to a constructed clinical dictionary and two model-optimized knowledge-enhancement metrics. The iterative framework is designed to progressively learn, starting from gaining a general understanding of the patient's condition based on raw reports and gradually refines and extracts critical information essential to the fine-grained analysis tasks. The effectiveness of the proposed framework is validated on various downstream medical image analysis tasks, including disease classification, region-of-interest segmentation, and phrase grounding. Our framework surpasses seven state-of-the-art methods in both fine-tuning and zero-shot settings, demonstrating its encouraging potential for different clinical applications.}

\maketitle

\begin{figure*}[htb]
\centering
\includegraphics[scale=.4]{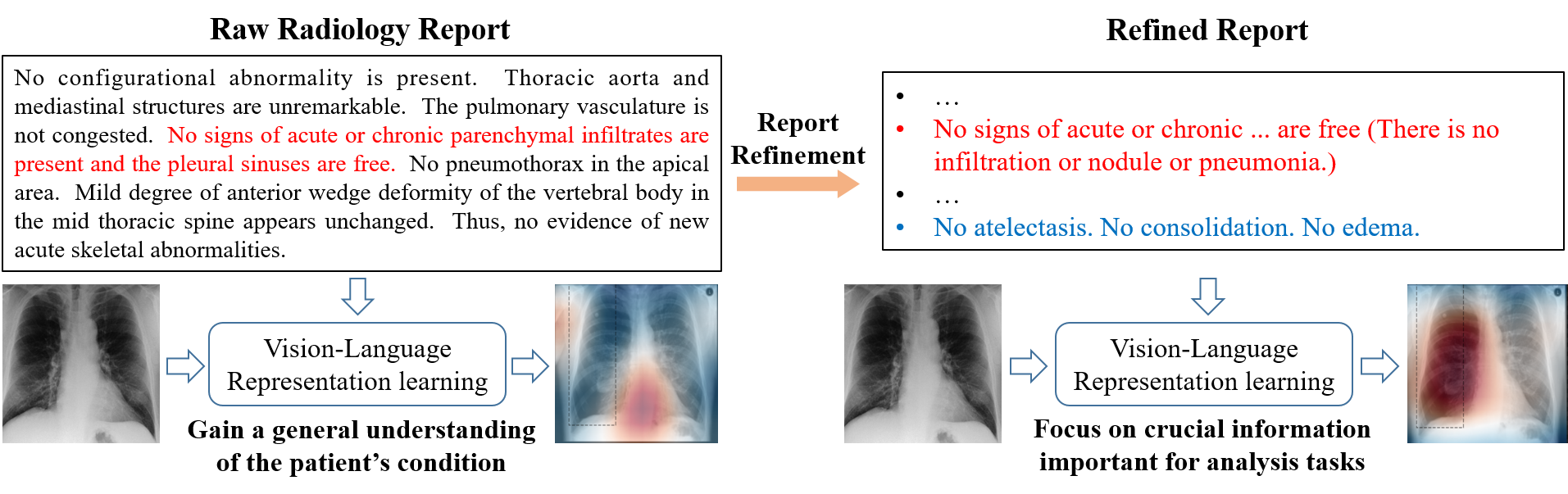}
\caption{Our proposed iterative vision-language representation learning framework. In the first iteration, the raw radiology reports are leveraged to gain a general understanding of the patient's condition. In the second stage, the refined reports are employed to further fine-tune the model, guiding the model towards capturing crucial information.}
\label{fig1}
\end{figure*}

\section{Introduction}
\label{sec1}
%First paragraph 引出label-efficient learning对于医学领域的重要性（annotation费时费力）
In modern clinical practice, medical imaging plays a crucial role in the detection, monitoring of progression, and evaluation of treatment prognosis for various diseases \citep{Mei2020,Huang2022,li2022artificial,Swanson2023}. However, the exponential growth of imaging data poses a significant burden to radiologists, impacting the efficiency of clinical workflows. To tackle this issue, artificial intelligence, particularly deep learning, has emerged as a revolutionary technique, automating medical image analysis and aiding in clinical decision-making \citep{Huang2021Registration, Zhou2021DUNet, Gu2023CDDSA, Wang2024AVDNet}. Nevertheless, manually annotating large datasets necessary to train respective deep learning models for each task is time-consuming and requires the expertise of domain specialists \citep{Wang2021AIDE,Xu2023}. As a result, there is an urgent need to develop effective medical foundation models that can handle various downstream tasks without relying on collecting large-scale labeled datasets \citep{Zhang2024foundation, Wang2023PyMIC}.

%Second paragraph 介绍Visual-language representation learning及存在的挑战
One promising and natural solution is to leverage the valuable information embedded within radiology reports \citep{Zhang2024foundation}, which are routinely collected in clinical practice. These reports contain rich domain knowledge that can effectively assist and guide image representation learning, thereby reducing the need for costly manual labels. A straightforward method in this direction involves extracting supervision signals directly from the reports. Various techniques, such as natural language processing (NLP) techniques and rule-based labelers, have been proposed for this purpose \citep{Wang2017, Irvin2019}. However, these labeling techniques often rely on fixed lexicons and manually engineered rules, making it difficult to adapt them to new scenarios. Another successful avenue is the application of implicit supervision through vision-language representation learning, which has demonstrated great success in natural image recognition tasks \citep{Wang2023, Kwon2023, Chen2023PaLi}. Nevertheless, the transfer of this technique to the medical domain faces many challenges \cite{Zhang2022ConVIRT}. One of the major obstacles is the presence of complex and sometimes redundant medical entity descriptions within radiology reports, which can pose significant difficulties for effective representation learning \citep{Wu2023MedKLIP, Zhang2023KAD}.  

%Third paragraph 介绍本工作
In this study, our primary objective is to enhance the medical vision-language foundation model by proposing a key semantic knowledge-emphasized report refinement method. Incorporating the proposed report refinement method, we develop a novel iterative vision-language representation learning framework. On the one hand, to refine the reports, we construct a simple yet effective clinical dictionary to link keywords in raw radiology reports with medical knowledge-supplemented sentences. Then, two model-optimized knowledge-enhancement metrics are constructed to guide the report refinement process using the medical knowledge-supplemented sentences such that the key information relevant to fine-grained downstream analysis tasks is effectively highlighted. On the other hand, our iterative framework enables the model to progressively learn the intricate medical information contained in radiology reports (Fig. \ref{fig1}). In the first iteration, we utilize the raw radiology reports as the initial source of information to gain a general understanding of the patient's condition, as provided by the radiologists. This step serves as a preliminary knowledge extraction process, obtaining the model to calculate the two model-optimized knowledge-enhancement metrics. In the second stage, we employ refined reports to further fine-tune the model, directing its attention towards crucial information.
The effectiveness of the proposed framework is validated by extensive downstream medical image analysis experiments, including disease classification, region-of-interest segmentation, and phrase grounding. The results demonstrate that our framework surpasses state-of-the-art vision-language representation learning methods in both fine-tuning and zero-shot settings.

%Fourth paragraph 总结main contributions
Our main contributions can be summarized as follows: 
\begin{itemize}
    \item We develop a novel iterative vision-language representation learning framework, which is designed to firstly gain a general understanding of the patient's condition from the raw radiology reports and then extract critical information by refining reports to capture the essential fine-grained features from the images. 
    \item We propose a key semantic knowledge-emphasized report refinement method. Under the guidance of a specially constructed clinical dictionary and two model-optimized knowledge-enhancement metrics, the reports are refined to highlight crucial information essential to fine-grained downstream image analysis tasks.
    \item Extensive experimental validations were conducted on multiple external datasets, covering various medical image analysis tasks. The results demonstrate that our proposed framework outperforms recent state-of-the-art vision-language representation methods in both fine-tuning and zero-shot settings, showcasing its effectiveness and robustness.
\end{itemize}

\section{Related work}
\label{sec2}
In this section, we provide a concise overview of recent research that focuses on utilizing information from radiology reports for medical image representation learning. We categorize these works into two main groups based on the strategies they employ: those that use explicit supervision signals extracted from radiology reports and those that incorporate implicit supervision through multi-modal vision-language representation learning.

\subsection{Report-supervised medical image representation learning}
\label{sec2.1}
Utilizing explicit supervision signals extracted from radiology reports to supervise the learning of medical imaging models is an intuitive and straightforward approach, particularly in scenarios where manual labels are not readily available. Wang et al. \cite{Wang2017} demonstrated the feasibility of this approach by constructing a chest X-ray dataset (ChestX-ray8) at a hospital scale. They employed NLP tools to search for the presence of 8 common thoracic pathology keywords in corresponding radiology reports and developed specific rules to remove negation and uncertainty. Then, they built a weakly supervised classification and localization framework using this dataset, validating the effectiveness of the automatically generated labels for these two important medical image analysis tasks. Another notable work in this area is by Irvin et al. \cite{Irvin2019}, who developed a rule-based labeler to extract structured labels for images from free-text radiology reports. Their efforts resulted in the construction of the well-known dataset, CheXpert. Leveraging the extracted labels, they trained convolutional neural networks using different uncertainty approaches to classify 14 observations, and their best model achieved higher performance than 3 additional radiologists on detecting 3 out of 5 selected pathologies, cardiomegaly, edema, and pleural effusion, validating the effectiveness of the generated labels for the detection of common chest radiographic observations.

Despite the effectiveness of these report-based automatic labeling approaches on individual datasets and specific tasks, there are two main limitations that hinder their widespread adoption. One limitation is that the relevant labeling rules are manually crafted. This manual process can introduce inaccuracies, resulting in the generation of incorrect labels \citep{Zhang2022ConVIRT}. Moreover, these labeling rules are often designed to capture a limited set of clinical observations mentioned in the reports, which can potentially overlook important information contained within the reports \citep{Zhang2022ConVIRT}. Another limitation of the approach is that the labeling rules are domain- and style-specific, and they rely on fixed lexicons. Consequently, the effectiveness of the developed labeling techniques may not generalize well to new scenarios or different datasets \citep{Zhou2023MRM}.

\begin{figure*}[htb]
\centering
\includegraphics[scale=.4]{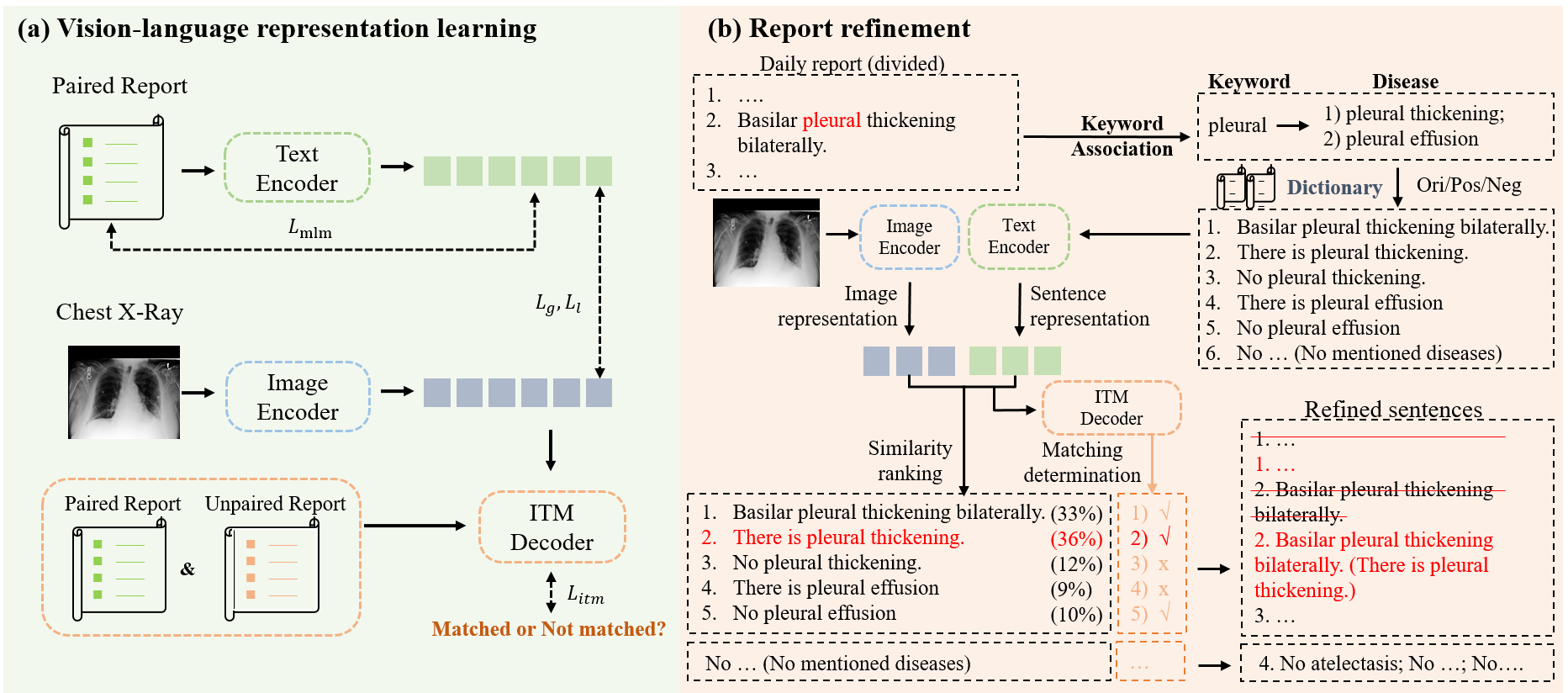}
\caption{The two major components of our framework. (a) The vision-language representation learning model with image-text matching determination capability. (b) The key semantic knowledge-emphasized report refinement method.}
\label{fig2}
\end{figure*}

\subsection{Medical vision-language representation learning}
\label{sec2.2}
Different from the approach of directly extracting supervision signals from radiology reports, medical vision-language representation learning leverages implicit supervision from the reports by simultaneously learning multi-modal representations. There are two main categories of methods in this area: those employing masked autoencoders \citep{Chen2022M3AE,Zhou2023MRM} and those employing contrastive learning techniques \citep{Zhang2022ConVIRT,isbiliu, Huang2021GLoRIA,Tiu2022CheXzero,isbihuang,Zhou2022REFERS,Liu2023MFLAG}. Methods using masked autoencoders aim to learn vision-language representations by restoring the original images and reports. For example, Chen et al. \cite{Chen2022M3AE} focuses on learning joint vision-language representations, which can be applied to various downstream tasks such as visual question answering, image-text classification, and image-caption retrieval. Zhou et al. \cite{Zhou2023MRM} targets the learning of radiolographic representations for disease diagnosis. These methods based on masked autoencoders typically require fine-tuning for evaluation due to the discrepancy between the pre-text restoration task and the downstream medical image recognition tasks. They may lack zero-shot capability. Contrastive learning-based methods, on the other hand, learn vision-language representations by aligning the distributions of multi-modal features. These methods can be employed for medical image recognition tasks in both fine-tuning and zero-shot settings. Among these works, Zhang et al. \cite{Zhang2022ConVIRT} introduced the pioneering framework ConVIRT, which learns medical image representations from paired images and reports by employing a bidirectional contrastive objective. Similarly, Tiu et al. \cite{Tiu2022CheXzero} proposed CheXzero, which achieved expert-level detection of pathologies without fine-tuning using labeled data. To capture localized features and fine-grained semantics in medical images, Huang et al. \cite{Huang2021GLoRIA} proposed GLoRIA, a multimodal global-local representation learning framework. Zhou et al. \cite{Zhou2022REFERS} introduced REFERS, which performs report generation in addition to multi-modal contrastive learning to facilitate the learning of well-transferable image representations. To simplify the training of medical vision-language representation models, Liu et al. \cite{Liu2023MFLAG} proposed M-FLAG, which trains only the vision model while freezing the language model.

While medical vision-language representation learning has shown promising results, there are still challenges to overcome. One major obstacle is the presence of complex and sometimes even redundant medical entity descriptions within radiology reports. To address this issue, researchers have explored different approaches. For example, Boecking et al. \cite{Boecking2022BioVil} trained a radiology-specific text encoder, called CXR-BERT, in their BioViL model to better handle radiology reports. Additionally, Wu et al. \cite{Wu2023MedKLIP} and Zhang et al. \cite{Zhang2023KAD} proposed methods to simplify radiology reports by extracting medical-related information before inputting them into the text encoders. The primary objective of our study is to refine the radiology reports by emphasizing key semantic knowledge to enhance vision-language representation learning. Different from the approaches of Wu et al. \cite{Wu2023MedKLIP} and Zhang et al. \cite{Zhang2023KAD}, which employ specific modules to extract medical entities from the reports, we propose a dictionary and model-dependent radiology report refinement method. Specifically, we develop an iterative vision-language representation learning framework. In the first iteration, we train a high-capacity vision-language representation learning model using the images and raw radiology reports. Then, we construct a clinical dictionary, which links keywords in raw radiology reports with medical knowledge-supplemented sentences. With the trained image and text encoders, we calculate two model-optimized knowledge-enhancement metrics to guide the report refinement process. In the second iteration, we replace the raw radiology reports with the refined versions to fine-tune the vision-language representation learning model. Importantly, we do not design specific modules to simplify the reports. Instead, we rely on the dictionary and the model itself to identify and focus on the important information. By accessing both the raw radiology reports and the refined versions, our method avoids the loss of critical details while making full use of the reports, benefiting from the hints provided by the refined knowledge.

\section{Method}
\label{sec3}
The primary objective of this study is to refine the radiology reports to enhance vision-language representation learning via a novel iterative learning framework. Here, we first describe the iterative vision-language representation learning framework in Sec.\ref{sec3.1}. Our main contribution of knowledge refinement for radiology reports is presented in Sec.\ref{sec3.2}.

\subsection{Iterative vision-language representation learning}
\label{sec3.1}
Our proposed framework involves training a multi-scale contrastive learning model in two iterations. In the first iteration, the model takes images and raw radiology reports as inputs, aiming to gain a general understanding of the patient's condition. Then, the radiology reports are refined using our designed key semantic knowledge-emphasized refinement method. It is important to note that this refinement method utilizes the model trained in the first iteration. In the second iteration, the inputs to the model are changed to images and refined reports. This iteration focuses on further refining the model to capture the essential fine-grained features from the images, which are crucial for downstream analysis tasks. The details of the multi-scale contrastive learning model (Fig. \ref{fig2}(a)) are described in this section, while the report refinement method (Fig. \ref{fig2}(b)) will be elaborated in Sec. \ref{sec3.2}.

\subsubsection{Multi-scale contrastive learning}
\label{sec3.1.1}
In this section, we introduce the multi-scale contrastive learning model, a simple but effective vision-language model comprising three components: the image encoder $E_{v}$ with the image projection block $P_{v}$, the text encoder $E_t$ with the text projection block $P_t$, and an ITM decoder.

We utilize a Vision Transformer (ViT) as our image encoder \citep{Dosovitskiy2021ViT}, which is adept at learning complex spatial relationships and capturing long-range dependencies within images. Let $\tilde{r_v} = E_{v}(X_v) \in \mathbb{R}^{K_v \times D_v}$ denote the extracted image features, where $K_v$ represents the number of image patches augmented by an additional [class] token \citep{Dosovitskiy2021ViT}, and $D_v$ is the dimension of the image features. The projection block $P_{v}$ then projects $\tilde{r_v}$ into a multi-modal joint space of $C$ dimensions, yielding local image representations ${r_v}=P_v(\tilde{r_v}) \in \mathbb{R}^{K_v \times C}$ for each input image patch. By applying average pooling to $r_v$ (except for the [class] embedding), we obtain the global image representations $R_v \in \mathbb{R}^{C}$.

For each radiology report, we first divide it into multiple sentences. A wordpiece tokenizer is then utilized to convert the processed reports into a sequence of tokens that can be analyzed by the text encoder $E_t$. We adopt BERT \citep{Devlin2019BERT} as our text encoder $E_t$. The text encoder generates the corresponding sentence representations as $\tilde{r_t} = E_{t}(X_t) \in \mathbb{R}^{K_t \times D_t}$, where $K_t$ indicates the number of divided sentences augmented by an additional [class] token, and $D_t$ is the dimension of the extracted text features. The projection block $P_t$ then projects $\tilde{r_t}$ into the joint space of $C$ dimensions. We achieve the global text representations $R_t \in \mathbb{R}^{C}$ by leveraging the [class] token embedding, while the remaining embeddings form the local text representations $r_t = P_t(\tilde{r_t}) \in \mathbb{R}^{K_t \times C}$.

A symmetric contrastive loss for the global alignment between the image and text representations is calculated:
\begin{equation}
\label{eq1}
\begin{split}
L_g = -\frac{1}{N}\sum_{i=1}^{N}(& log\frac{exp(<R_{vi}, R_{ti}>/\tau_1)}{\sum_{j=1}^{N}exp(<R_{vi}, R_{tj}>\tau_1)} \\& +log\frac{exp(<R_{ti}, R_{vi}>/\tau_1)}{\sum_{j=1}^{N}exp(<R_{ti}, R_{vj}>\tau_1)})
\end{split}
\end{equation}
$<R_{vi}, R_{ti}>$ indicates the cosine similarity between the two global features $R_{vi}$ and $R_{ti}$. $N$ is the batch size. $\tau_1$ is a temperature parameter.

Inspired by GLoRIA \citep{Huang2021GLoRIA}, we propose a local contrastive loss to promote fine-grained alignment between images and reports. Unlike GLoRIA, we choose sentences sampled from the original reports as the basic contrastive units instead of individual words to capture complete semantic representations. Specifically, a similarity matrix $s = r_t \cdot r_v^T \in \mathbb{R}^{K_t \times K_v}$ is obtained by computing the dot product of the local text and image representations, which reflects the similarity between sentence representations and image patch representations. Here, $s_{i,j}$ corresponds to the similarity between sentence $i$ in the text and patch $j$ in the image. The matrix is then normalized as $a_{i,j} = \frac{exp(s_{i,j}/\tau_2)}{\sum_{k=1}^{K_v}exp(s_{i,k}/\tau_2)}$, where $\tau_2$ is a temperature parameter. Similar to GLoRIA, we derive the context-enhanced image representations for each given sentence as $\hat{r_{vi}} =\sum_{j=0}^{K_v}a_{i,j}r_{vj}$. Finally, we aggregate the similarities between all $K_t$ sentences and their corresponding enhanced image representations using a matching function $M$, defined as $M(X_v, X_t) = \log(\sum_{i=1}^{K_t}exp(<\hat{r_{vi}}, r_{ti}>/\tau_3))^{\tau_3}$, where $\tau_3$ is also a temperature parameter. The symmetric local contrastive loss is then defined as:
\begin{equation}
\label{eq2}
\begin{split}
L_l = -\frac{1}{N}\sum_{i=1}^{N}(&log\frac{exp(M(X_{vi}, X_{ti})/\tau_2)}{\sum_{j=1}^{N}exp(M(X_{vi}, X_{tj})/\tau_2)} \\&+log\frac{exp(M(X_{vi}, X_{ti})/\tau_2)}{\sum_{j=1}^{N}exp(M(X_{vj}, X_{ti})/\tau_2)})
\end{split}
\end{equation}

\subsubsection{Image-text matching determination}
\label{sec3.1.2}
Image-Text Matching (ITM) is a binary classification task in which the model utilizes a language decoder with a linear layer to predict whether the input image-text pairs are positive (matching) or negative (non-matching). The primary objective of ITM is to capture fine-grained alignment between visual and textual information \citep{li2021align}. In this paper, we introduce an improved ITM mainly to determine the correctness of sentences generated from a dictionary (Sect. \ref{sec3.2.2}). For implementation, we employ a BERT base language decoder. It should be noted that we deviate from the classical ITM approach \citep{li2022blip} by not incorporating the hard-sample strategy, as it is not suitable in the context of medical reports where text with similar semantics is frequently present. Instead, we randomly select one sentence from an unpaired report to serve as the negative sample.

We employ a straightforward binary cross-entropy loss for training the ITM module, denoted as $L_{itm}$. Finally, in combination with the masked language model loss $L_{mlm}$ utilized in $E_t$, the loss used for training the multi-scale contrastive learning model is:

\begin{equation}
\label{eq3}
L =  L_g + L_l +L_{itm} + L_{mlm} 
\end{equation}

\subsection{Key semantic knowledge-emphasized report refinement}
\label{sec3.2}
Following the preparation outlined in Sec \ref{sec3.1}, we have a pre-trained model with image-text matching determination capability ready to extract preliminary visual-language features. In this section, we focus on report refinement for fine-tuning the pre-trained model, as depicted in Fig. \ref{fig2}(b).

\subsubsection{Construction of the clinical dictionary}
\label{sec3.2.1}
To facilitate report refinement, we start by constructing a clinical dictionary. This dictionary is utilized to extract keywords from complex sentences and supplement the sentences with relevant knowledge-emphasized sentences, aiding in semantic understanding. The construction of the dictionary begins with the identification of diseases. In total, we include 17 diseases, out of which 14 diseases are adopted from the NIH ChestX-ray dataset, while the remaining 3 diseases are determined manually based on their frequency of occurrence in the MIMIC-CXR V2 dataset. Each disease is then linked to one or several keywords. For example, the keyword for the disease ``atelectasis'' is simply ``atelectasis''. In cases where diseases can be described differently within the MIMIC-CXR V2 dataset, we add additional relevant keywords to ensure comprehensive coverage. For instance, the keywords for the disease ``pneumothorax'' are ``pneumothor, air, chest wall, acute cardiopulmonary''. 

Utilizing the identified diseases, we manually add one positive sentence for each disease in the format ``There is $\left\{\_\right\}$" where ``$\left\{\_\right\}$'' represents the disease, and one negative sentence in the format ``No $\left\{\_\right\}$''.  Additionally, we designed a prompt to generate descriptive sentences of the diseases using large language models, ``Claude-3-Sonnet'' and ``GPT-4o'', to explore the potential of these models in our methodology. The prompt for positive sentences is: ``According to the diseases listed by me, please give five possible imaging diagnosis reports for each disease, which can give specific possible parts and corresponding image descriptions, and different diseases should be differentiated. Answer in English, be professional, concise, short, and as comprehensive as possible. Start the description directly, without `chest X-ray' and other statements, no numbers. lung opacity, atelectasis, cardiomegaly, infiltration, mass, nodule, pneumonia, pneumothorax, consolidation, edema, emphysema, fibrosis, pleural thickening, hernia, pleural effusion, rib fractures, embolism." For negative sentences, we add the term ``negative'' in the prompt before ``image diagnosis reports''. In the supplementary file, we provide the details of the constructed clinical dictionaries (Appendix Table S1 and Table S2).

\subsubsection{Report refinement}
\label{sec3.2.2}
Using the constructed clinical dictionary, we establish keyword associations between each sentence in the radiology report and the keywords listed in the dictionary. When a sentence is found to be related to a specific disease type, we supplement the original sentence with the generated positive and negative sentences related to that disease. For example, in the sentence ``It appears more amorphous and a small focus of infection is not excluded.'', the keyword ``infection'' of the disease ``pneumonia'' is detected. In this case, we temporarily supplement this sentence with the 6 positive sentences (``Lobar consolidation in the right lower lobe, consistent with bacterial pneumonia.'' ... ``There is pneumonia.'') and 6 negative sentences (``No radiographic evidence of pneumonia or consolidation.'' ... ``No pneumonia.'') generated for the disease ``pneumonia''. If the detected keyword is linked to multiple diseases (e.g., ``pleural'' is linked to both ``pleural thickening'' and ``pleural effusion''), all the sentences generated for the different diseases are included for consideration.

Subsequently, we feed the image, along with all the sentences, including the original sentence and the supplemented ones, into the trained contrastive model and ITM decoder. The outputs are then utilized to assess whether the supplemented sentences accurately describe the patient's condition. We employ two metrics for this assessment (Fig. \ref{fig2}(b)): 1) a similarity metric determined based on the pre-trained contrastive model, and 2) a matching metric obtained using the trained ITM decoder with a binary linear classifier. The evaluation of the supplemented sentences is conducted in two steps. In the first step, we determine the priority of the different sentence groups. For example, if a sentence is detected with the keyword ``pleural'', it will have five sentence groups: the original sentence (G1), 6 positive sentences of the disease ``pleural thickening'' (G2), 6 negative sentences of the disease ``pleural thickening'' (G3), 6 positive sentences of the disease ``pleural effusion'' (G4), and 6 negative sentences of the disease ``pleural effusion'' (G5). For each sentence, we obtain its similarity score. Then, the average similarity score ($ss$) for each sentence group is calculated to determine the priority. Assuming the average similarity scores of the five groups are as follows: $ss_{G1}=30\%$, $ss_{G2}=32\%$, $ss_{G3}=15\%$, $ss_{G4}=25\%$, $ss_{G5}=10\%$. Then, the sentences are sorted according to the following order: G2, G1, G4, G3, G5. Within each group, the different sentences are sorted according to their respective similarity scores, with the ones having larger similarity scores placed before the ones with smaller similarity scores.

After sorting the sentences, we proceed to select the most suitable generated sentence to supplement the original sentence in the report, taking into account both the similarity metric and the matching metric. Specifically, we examine the manually generated sentence in each sentence group to determine whether the sentences in the group (describing the presence or absence of disease) can match the image through ITM. If ITM indicates a match between the sentence group and the image, we have the option to choose either the sentence with the highest similarity score or the manually generated sentence from that group to supplement the original sentence in the radiology group. In the results section, we will demonstrate that the latter approach is more effective for our implementation. If all the groups listed before the original sentence do not match the image, we retain the original sentence without any generated sentence supplementation. However, if the original sentence is listed as the first sentence group, we only assess the matching performance of the second sentence group to determine whether the original sentence will be supplemented with a generated sentence from that group or not. After reviewing all the sentences in the report, we take an additional step to supplement the report. We add three manually generated negative sentences with the highest similarity scores for diseases that are not detected in the report. These sentences serve to provide additional information by explicitly stating the absence of these diseases.

Finally, we utilize the refined reports enriched with newly acquired knowledge from the supplemented sentences for fine-tuning the model, directing the model's attention towards crucial information.

\section{Experiments and results}
\label{sec4}

\begin{table*}[h]
\begin{center}
\caption{Fine-tuning disease classification results (AUC: \%) on three datasets using varying ratios of annotations for fine-tuning. The symbol ``-'' indicates that no results have been reported in the existing literature for the corresponding experimental setting. The symbol ``$\dagger$'' denotes the lack of zero-shot capabilities, whereas ``$\ddagger$'' signifies the use of classification labels during the pre-training phase. The top-2 results are highlighted in bold.}
\label{tab1}
\begin{tabular}{c|c|c|c|c|c|c|c|c|c}
\hline
\multirow{2}{*}{Methods} & \multicolumn{3}{|c|}{RSNA Pneumonia} & \multicolumn{3}{|c|}{CheXpert} & \multicolumn{3}{|c}{NIH ChestX-ray} \\
% \cline{2-10}
~   & 1\% & 10\% & 100\% & 1\% & 10\% & 100\% & 1\% & 10\% & 100\% \\
\hline
REFERS$^\dagger$                       & 89.4 & 91.6 & 92.7  & 87.2 & 88.1 & 88.2& 76.7 & 80.9 & 84.7  \\
M3AE$^\dagger$                         & 89.0 & 90.8 & 92.3  & 86.2 & 87.3 & 87.9& - & - & -\\
MRM$^\dagger$                          & \textbf{91.3} & \textbf{92.7} & \textbf{93.3} & 88.5 & 88.5 & 88.7 & \textbf{79.4} & \textbf{84.0} & \textbf{85.9}\\
MGCA$^\ddagger$\citep{Liu2023MFLAG}    & 89.1 & 89.9 & 90.8  & \textbf{88.8} & \textbf{89.1} & \textbf{89.7}& 61.1 & 67.8 & 77.3 \\ \hline
ConVIRT \citep{Liu2023MFLAG}           & 77.4 & 80.1 & 81.3  & 85.9 & 86.8 & 87.3& 60.0 & 69.0 & 76.6 \\
M-FLAG                                & - & - & -  & 64.4 & 71.4 & 78.1& 62.2 & 71.6 & 78.7 \\
GLoRIA \citep{Liu2023MFLAG}            & 86.1 & 88.0 & 88.6 & 86.6 & 87.8 & 88.1& 60.1 & 71.2 & 77.7  \\
BioViL \citep{Wu2023MedKLIP}           & 88.1 & 88.4 & 89.1 & - & - & - & 69.5   & 75.3 & 82.5 \\
MedKLIP \citep{Wan2023MedUniC}         & 87.3 & 88.0 & 89.3  & 86.2  & 86.5  & 87.7 & 77.2 & 78.9 & 83.2\\
Ours                                  & \textbf{91.4} & \textbf{92.4} & \textbf{93.4} & \textbf{88.7} & \textbf{89.0} & \textbf{89.0}& \textbf{78.7} &\textbf{83.5} & \textbf{85.7}  \\

\hline
% \bottomrule
\end{tabular}
\end{center}
%\resizebox{\linewidth}{!}{
\end{table*}

\begin{table*}[h]
\tiny
% \scriptsize
\centering
\caption{Disease-level classification results (AUC: \%) on the NIH ChestX-ray dataset using different ratios of labeled samples for fine-tuning.}
\label{tab2}
\setlength{\tabcolsep}{2.0pt}

\begin{tabular}{c|c|ccccccccccccccc}
\hline
\multicolumn{1}{c|}{\rotatebox{90}{Labeling Ratios} } &
Methods &
\rotatebox{90}{Average} &
\rotatebox{90}{Atelectasis} &
\rotatebox{90}{Cardiomegaly} &
\rotatebox{90}{Consolidation} &
\rotatebox{90}{Edema} &
\rotatebox{90}{Effusion} &
\rotatebox{90}{Emphysema} &
\rotatebox{90}{Fibrosis} &
\rotatebox{90}{Hernia} &
\rotatebox{90}{Infiltration} &
\rotatebox{90}{Mass} &
\rotatebox{90}{Nodule} &
\rotatebox{90}{Pleural Thickening} &
\rotatebox{90}{Pneumonia} &
\rotatebox{90}{Pneumothorax} \\ \hline
  \multicolumn{1}{c|}{\multirow{7}{*}{1\%}} &
REFERS &76.7  &77.5 &85.6 &78.6 &84.9 &85.4 &79.5 &\textbf{72.3} &77.1 &67.5&76.2 &66.5 &\textbf{71.6} &69.3 &81.7\\
\multicolumn{1}{c|}{} &
  Model Genesis &  70.3 &  72.1 &  67.1 &  75.8 &  76.1 &  80.6 &  72.6 &  64.8 &  73.5 &  65.7 &  65.2 &  62.2 &  67.6 &  64.8 &  76.2 \\
\multicolumn{1}{c|}{} &
  C2L &  71.1 &  75.1 &  67.1 &  77.6 &  75.1 &  83.4 &  71.5 &  66.8 &  70.0 &  63.8 &  70.1 &  66.2 &  68.1 &  65.7 &  74.4 \\
\multicolumn{1}{c|}{} &
  Context Restoration &  67.8 &  69.1 &  64.4 &  73.2 &  73.8 &  78.1 &  70.0 &  62.1 &  70.2 &  65.2 &  62.4 &  59.1 &  65.0 &  62.2 &  73.8 \\
\multicolumn{1}{c|}{} &
  TransVW &
  71.3 &  74.5 &  68.9 &  76.7 &  79.8 &  81.1 &  67.9 &  68.7 &  68.2 &  66.8 &  66.5 &  66.2 &  68.5 &  68.8 &  75.0 \\
\multicolumn{1}{c|}{} &
  ImageNet Pre-training &
  69.8 &  73.3 &  69.6 &  76.0 &  81.7 &  80.5 &  67.1 &  64.9 &  64.8 &  65.8 &  67.0 &  62.3 &  65.7 &  65.0 &  74.0 \\
\multicolumn{1}{c|}{} &
Ours &\textbf{78.7} &\textbf{77.8}&\textbf{90.0}&\textbf{79.5}&\textbf{87.1}&\textbf{86.8}&\textbf{86.0}&71.7&\textbf{82.3}&\textbf{67.8}&\textbf{81.3}&\textbf{67.5}&69.8&\textbf{69.5}&\textbf{83.9}  \\ 
  \hline

  \multicolumn{1}{c|}{\multirow{7}{*}{10\%}} &
REFERS &80.9 &80.1 &89.8 &79.5 &87.8 &87.5 &88.2 &77.2 &86.1 &69.6 &82.0 &72.8 &74.2 &72.2 &85.6\\
\multicolumn{1}{c|}{} &
  Model Genesis &
  76.0 &  77.2 &  72.8 &  77.5 &  85.7 &  85.2 &  81.0 &  75.3 &  78.0 &  68.4 &  73.1 &  69.5 &  72.2 &  67.7 &  80.4 \\
\multicolumn{1}{c|}{} &
  C2L &  76.6 &  78.0 &  75.5 &  77.5 &  84.1 &  85.7 &  81.2 &  73.7 &  79.5 &  67.4 &  77.5 &  71.7 &  72.0 &  67.3 &  81.9 \\
\multicolumn{1}{c|}{} &
  Context Restoration &
  73.8 &  75.5 &  70.6 &  77.1 &  84.5 &  84.2 &  79.4 &  73.1 &  67.5 &  68.1 &  70.9 &  66.9 &  71.7 &  65.2 &  79.1 \\
\multicolumn{1}{c|}{} &
  TransVW &
  74.4 &  76.5 &  70.8 &  77.6 &  83.0 &  84.8 &  79.7 &  69.9 &  74.7 &  68.5 &  72.1 &  68.3 &  72.4 &  63.2 &  79.6 \\
\multicolumn{1}{c|}{} &
  ImageNet Pre-training &
  74.4 &  74.2 &  79.8 &  75.9 &  85.7 &  83.2 &  80.4 &  72.1 &  74.0 &  64.1 &  71.7 &  65.6 &  69.6 &  66.2 &  79.7 \\

\multicolumn{1}{c|}{} &
Ours &\textbf{83.5} & \textbf{82.1} &\textbf{91.8}&\textbf{80.9}&\textbf{88.8}&\textbf{88.5}&\textbf{91.2}&\textbf{83.1}&\textbf{94.4}&\textbf{70.4}&\textbf{85.9}&\textbf{74.5}&\textbf{76.4}&\textbf{73.6}&\textbf{87.5}  \\ 
  \hline

  \multicolumn{1}{c|}{\multirow{7}{*}{100\%}} &
REFERS &84.7 &83.0 &92.3 &82.1 &90.2 &88.7 &91.4 &83.9 &93.3 &\textbf{74.1} &85.5 &76.7 &78.5 &77.0 &89.1 \\
\multicolumn{1}{c|}{} &
  Model Genesis &  81.0 &  78.8 &  84.5 &  79.2 &  87.8 &  86.6 &  89.7 &  81.0 &  85.2 &  71.1 &  81.9 &  73.2 &  75.8 &  73.0 & 85.6 \\
\multicolumn{1}{c|}{} &
  C2L &
  82.2 &  81.1 &  90.2 &  81.0 &  88.1 &  88.0 &  88.3 &  80.8 &  86.8 &  72.0 &  82.7 &  74.1 &  76.2 &  75.3 &  85.9 \\
\multicolumn{1}{c|}{} &
  Context Restoration &
  78.7 &  75.8 &  82.9 &  76.4 &  86.6 &  84.8 &  88.2 &  78.6 &  83.0 &  70.0 &  79.6 &  69.5 &  73.2 &  69.4 &  84.0 \\
\multicolumn{1}{c|}{} &
  TransVW &
  81.7 &  79.8 &  85.0 &  80.0 &  88.2 &  87.1 &  90.1 &  81.8 &  85.9 &  72.3 &  82.6 &  74.4 &  76.6 &  74.0 &  86.1 \\
\multicolumn{1}{c|}{} &
  ImageNet Pre-training &
  80.0 &  78.3 &  89.3 &  77.6 &  87.9 &  85.9 &  87.4 &  78.5 &  88.8 &  65.9 &  79.9 &  70.7 &  74.5 &  71.0 &  84.7 \\

\multicolumn{1}{c|}{} &

Ours &\textbf{85.7} &\textbf{83.8}  &\textbf{92.6}  &\textbf{82.2}   &\textbf{90.6}   &\textbf{89.3}  &\textbf{93.6}  &\textbf{85.8}  &\textbf{96.0}  &72.3  &\textbf{87.5}  &\textbf{78.4} &\textbf{79.5}  &\textbf{78.2}  &\textbf{89.6}  \\ 
  \hline
\end{tabular}
\end{table*}

\subsection{Datasets}
\label{sec4.1}
For model training, we utilized the X-ray images and corresponding radiology reports from the MIMIC-CXR V2 dataset \citep{Johnson2019MIMIC}. The model evaluation involved four tasks: fine-tuning disease classification, fine-tuning region-of-interest segmentation, zero-shot disease classification, and zero-shot phrase grounding. To perform disease classification, we incorporated four X-ray datasets, CheXpert \citep{Irvin2019}, NIH ChestX-ray \citep{Wang2017}, RSNA Pneumonia \citep{Shih2019RSNA}, and SIIM-ACR Pneumothorax \footnote{https://www.kaggle.com/c/siim-acr-pneumothorax-segmentation}. For fine-tuning segmentation, we adopted the SIIM-ACR Pneumothorax dataset. For zero-shot phrase grounding, we utilized the MS-CXR dataset \citep{Boecking2022BioVil}.

MIMIC-CXR V2 is a large dataset comprising 377,110 chest X-rays obtained from 227,827 imaging studies \citep{Johnson2019MIMIC}. These studies were conducted at the Beth Israel Deaconess Medical Center between 2011 and 2016. Each image is accompanied by a corresponding free-text radiology report. In our study, we utilized all the available data from this dataset for model training.

CheXpert consists of 224,316 chest X-rays obtained from 65,240 patients. Following the official practice outlined by Irvin et al. \cite{Irvin2019}, we report the classification results for five selected pathologies, including atelectasis, cardiomegaly, consolidation, edema, and pleural effusion. Since the official test set is not publicly available, we adopted the approach from a previous study \citep{Zhang2022ConVIRT} and employed the official validation set as our test set. Additionally, we randomly sampled 5,000 samples from the official training set to construct our validation set, similar to Zhou et al. \cite{Zhou2023MRM}. For the fine-tuning evaluation, our training, validation, and testing sets contain 218,414, 5,000, and 234 images, respectively. For the zero-shot evaluation, only the 234 testing images were employed.

NIH ChestX-ray offers 112,120 chest X-rays for the classification of 14 pathologies \citep{Wang2017}. Similarly, for fine-tuning evaluation, we divided the dataset into training, validation, and testing sets, following a split ratio of 7:2:1.

RSNA Pneumonia provides data for the binary classification task of pneumonia vs. normal. In accordance with the official configuration, the training, validation, and testing sets consist of 25,184, 1,500, and 3,000 images, respectively \citep{Shih2019RSNA}.

SIIM-ACR Pneumothorax is a dataset that consists of more than 120,000 chest X-rays, with each image being accompanied by precise manual segmentation masks of the pneumothorax regions. Following Huang et al. \cite{Huang2021GLoRIA}, we divided the dataset into three subsets: 70\% for training, 15\% for validation, and 15\% for testing.

MS-CXR provides bounding box annotations along with paired sentences that describe the clinical findings \citep{Boecking2022BioVil}. The dataset includes a total of 1,162 annotations of 881 cases, and we utilized all of them for the evaluation.

\subsection{Comparison methods}
\label{sec4.2}
We compared our framework with various existing state-of-the-art vision-language representation learning methods to validate its effectiveness. These methods include ConVIRT \citep{Zhang2022ConVIRT}, GLoRIA \citep{Huang2021GLoRIA}, BioViL \citep{Boecking2022BioVil}, M3AE \citep{Kwon2023M3AE}, REFERS \citep{Zhou2022REFERS}, MGCA \citep{Wang2022MGCA}, MRM\citep{Zhou2023MRM}, CheXzero \citep{Tiu2022CheXzero}, MedKLIP \citep{Wu2023MedKLIP}, M-FLAG \citep{Liu2023MFLAG}, and Med-UniC \citep{Wan2023MedUniC}.  The specific contributions of most of these methods to the field have been discussed in Sec. \ref{sec2}. It should be noted that we have tried our best to supplement the missing results of these methods for different tasks. This was accomplished by either referring to other published literature \citep{Liu2023MFLAG, Wu2023MedKLIP, Wan2023MedUniC} or reproducing the experiments by ourselves.

Specifically, for the evaluation of fine-tuning disease classification, we compared our method with ConVIRT, GLoRIA, BioViL, M3AE, REFERS, MedKLIP, MGCA, MRM, and M-FLAG. For the evaluation of fine-tuning pneumothorax region segmentation, we compared our method with ConVIRT, GLoRIA, MGCA, M-FLAG, and Med-UniC. For the evaluation of zero-shot disease classification, we compared our method with ConVIRT, GLoRIA, BioViL, and CheXzero. For the zero-shot phrase grounding evaluation, we compared our method with ConVIRT, GLoRIA, and BioViL.

\subsection{Implementation details}
\label{sec4.3}
In our experiments, we adopted the widely used ViT-B/16 as our image encoder and BERT with a width of 768 as our text encoder and ITM decoder. We set the batch size to 128 and utilized the AdamW optimizer with a weight decay of 0.05, $\beta_1=0.9$, and $\beta_2=0.95$. For the first iteration, the mask ratio used in BERT was set to 0.15, and the initial learning rate was set to 1.5e-4 with 50 training epochs. For the second iteration, the mask ratio was set to 0, and the learning rate was adjusted to 3e-6 with 10 training epochs. All experiments were implemented using PyTorch, and we used four NVIDIA A100 GPUs in parallel.

For the fine-tuning disease classification evaluation, we utilized the SGD optimizer with a momentum of 0.9. The best learning rate was searched in the range of 1e-5 to 8e-3 to achieve optimal validation performance. For the fine-tuning segmentation task, we utilized the AdamW optimizer with the learning rate of 4e-6, 2.5e-5, and 2e-5 for 1$\%$, 10$\%$, and 100$\%$ labeling rations.

\subsection{Evaluation metrics}
\label{sec4.4}
For disease classification, we evaluated the performance using the area under the curve (AUC) score. We calculated the average AUC scores for the respective datasets, as well as disease-level scores whenever possible. For region-of-interest segmentation, we employed the Dice similarity coefficient (Dice). For phrase grounding, we assessed the results based on the Intersection over Union (IoU) score and the contrast-to-noise ratio (CNR) value. IoU measures the overlap between the generated saliency maps and the ground-truth segmentation labels. CNR evaluates the contrast difference inside and outside the bounding box.
%check check

\subsection{Results for fine-tuning disease classification}
\label{sec4.5}
In this section, we present the disease classification results of different methods in fine-tuning settings. Three datasets (CheXpert, NIH ChestX-ray, and RSNA Pneumonia) were evaluated. For each dataset, we utilized three percentages of labeled data (1\%, 10\%, and 100\%) during the fine-tuning process to investigate the influence of the fine-tuning sample number on the classification performance.

The AUC scores of different methods using varying ratios of annotations are shown in Table \ref{tab1}. For a more comprehensive comparison, methods that use classification labels (``$\ddagger$'') or lack zero-shot capabilities (``$\dagger$'') are also included. Our proposed method outperforms five state-of-the-art methods, including ConVIRT, M-FLAG, GLoRIA, BioViL, and MedKLIP, across three datasets. For instance, compared to MedKLIP, our method achieves a 4.1\%, 4.4\% and 4.1\% improvement on the RSNA Pneumonia dataset with 1\%,10\% and 100\% fine-tuning data, respectively. When compared with two leading methods in CXR fine-tuning classification tasks, MGCA (relying on classification annotations \citep{Zhou2023MRM, Wan2023MedUniC}) and MRM (lacking zero-shot capability \citep{zhao2023clip, he2024foundation}), our method demonstrates competitive performance. Generally speaking, the proposed method demonstrates comparable performance to existing algorithms in fine-tuning classification tasks, while also offering advantages in clinical scenarios with limited annotations (annotation-free) and task generalization (zero-shot capability).

In addition to the mean AUC scores, we also provide the classification results for the 14 chest pathologies specific to the NIH ChestX-ray dataset (Table \ref{tab2}). Here, we included several image self-supervised methods (Model Genesis \citep{Zhou2021ModelGenesis}, C2L \citep{Zhou2020C2L}, Context Restoration \citep{Chen2019ContextRes}, and TransVW \citep{HaghighiTransVW}), as well as an ImageNet pre-trained model \citep{Wang2017}, for comparison. Across most diseases, our proposed framework consistently achieves the highest AUC scores, thereby validating its effectiveness in fine-grained disease classification.

\begin{table}[h]
\centering
\caption{Results (Dice: \%) for pneumothorax region segmentation on the SIIM-ACR Pneumothorax dataset using different ratios of labeled samples for fine-tuning.}
\label{tab3}
\begin{tabular}{ccccc}
\hline
Methods  & 1\%  & 10\% & 100\% \\ \hline
ConVIRT  & 25.0 & 43.2 & 59.9  \\
GLoRIA   & 35.8 & 46.9 & 63.4  \\
MGCA     & 49.7 & 59.3 & 64.2  \\
M-FLAG   & 52.5 & 61.2 & 64.8  \\
Med-UniC & 56.7 & 62.2 & 64.4  \\
Ours &\textbf{61.3}  &\textbf{72.2}  & \textbf{88.7}\\ \hline
\end{tabular}
\end{table}

\subsection{Results for fine-tuning pneumothorax region segmentation}
\label{sec4.6}
Region-of-interest segmentation is another important medical image analysis task that offers various clinical applications, such as disease progression evaluation and treatment planning. We evaluate the effectiveness of our proposed framework by applying it to segment the pneumothorax regions using the SIIM-ACR Pneumothorax dataset in the fine-tuning setting. Results are reported in Table \ref{tab3}.

Similar to the fine-tuning disease classification task, three different ratios of labeled samples were employed to fine-tune the model for pneumothorax region segmentation. Significant improvements in segmentation performance across all three ratios are obtained. Notably, as more labeled samples are introduced for fine-tuning, the segmentation performance displays a more pronounced enhancement. Specifically, compared to the respective best-performing comparison methods, the Dice similarity score is increased by 4.6\%, 10\%, and 23.9\% at the labeling ratio of 1\%, 10\%, and 100\%, respectively.
Interestingly, this trend differs from what we observed in the fine-tuning disease classification task, where larger improvements were observed with fewer fine-tuning samples. One possible explanation for this discrepancy is that segmentation is a more challenging task than classification, demanding more localized representations. Our proposed framework learns strong visual representations, which can be further refined when provided with additional hints from the fine-tuning samples. Consequently, these refined representations enable better discriminatation of the target regions from the surrounding background. Nevertheless, it is an intriguing observation, and we plan to investigate it further in our following studies.

\begin{table}[h]
\centering
\caption{Zero-shot disease classification results (AUC: \%) on the RSNA Pneumonia and SIIM-ACR Pneumothorax datasets.}
\label{tab4}
\begin{tabular}{ccc}
\hline
\textbf{Methods}                                     & \textbf{RSNA} & \textbf{SIIM} \\ \hline
ConVIRT                                             & 80.4        & 64.3        \\
GLoRIA                                              & 71.5        & 53.4        \\
BioViL                                              & 82.8      & 70.8        \\
CheXzero                                            & 85.8       & 68.8        \\
Ours                                       & \textbf{87.7}       & \textbf{88.5}        \\ \hline
\end{tabular}
\end{table}

\subsection{Results for zero-shot disease classification}
\label{sec4.7}
In this section, we present the zero-shot disease classification results on two datasets, RSNA Pneumonia and SIIM-ACR Pneumothorax. Table \ref{tab4} lists the results obtained by different methods. Among the four comparison methods, different trends are observed in the AUC scores on the two datasets. On the RSNA Pneumonia dataset, GLoRIA achieves the lowest mean AUC score, while CheXzero obtains the highest score. On the SIIM-ACR Pneumothorax dataset, although GLoRIA consistently achieves the lowest classification results, the best score is given by BioViL instead of CheXzero. This discrepancy suggests that the two datasets may possess different data characteristics that affect the learning process.

Compared to the four existing state-of-the-art methods, our proposed framework consistently demonstrates enhanced performance on both datasets. Particularly, on the SIIM-ACR Pneumothorax dataset, our framework achieves a remarkable 17.7\% increase in AUC score when compared to the best-performing comparison method, BioViL, thereby validating the effectiveness of the proposed report refinement method and iterative learning process.

\begin{table}[h]
\centering
\caption{Zero-shot phrase grounding results on the MS-CXR dataset.}
\label{tab5}
\begin{tabular}{ccc}
\hline
Methods & CNR  & mIoU   \\ \hline
ConVIRT & 0.818 & 0.238  \\
GLoRIA  & 0.930 & 0.246 \\
BioViL  & 1.027 & 0.266 \\
Ours & \textbf{1.257} & \textbf{0.266} \\ \hline
\end{tabular}
\end{table}

\begin{figure*}[htb]
\centering
\includegraphics[scale=.4]{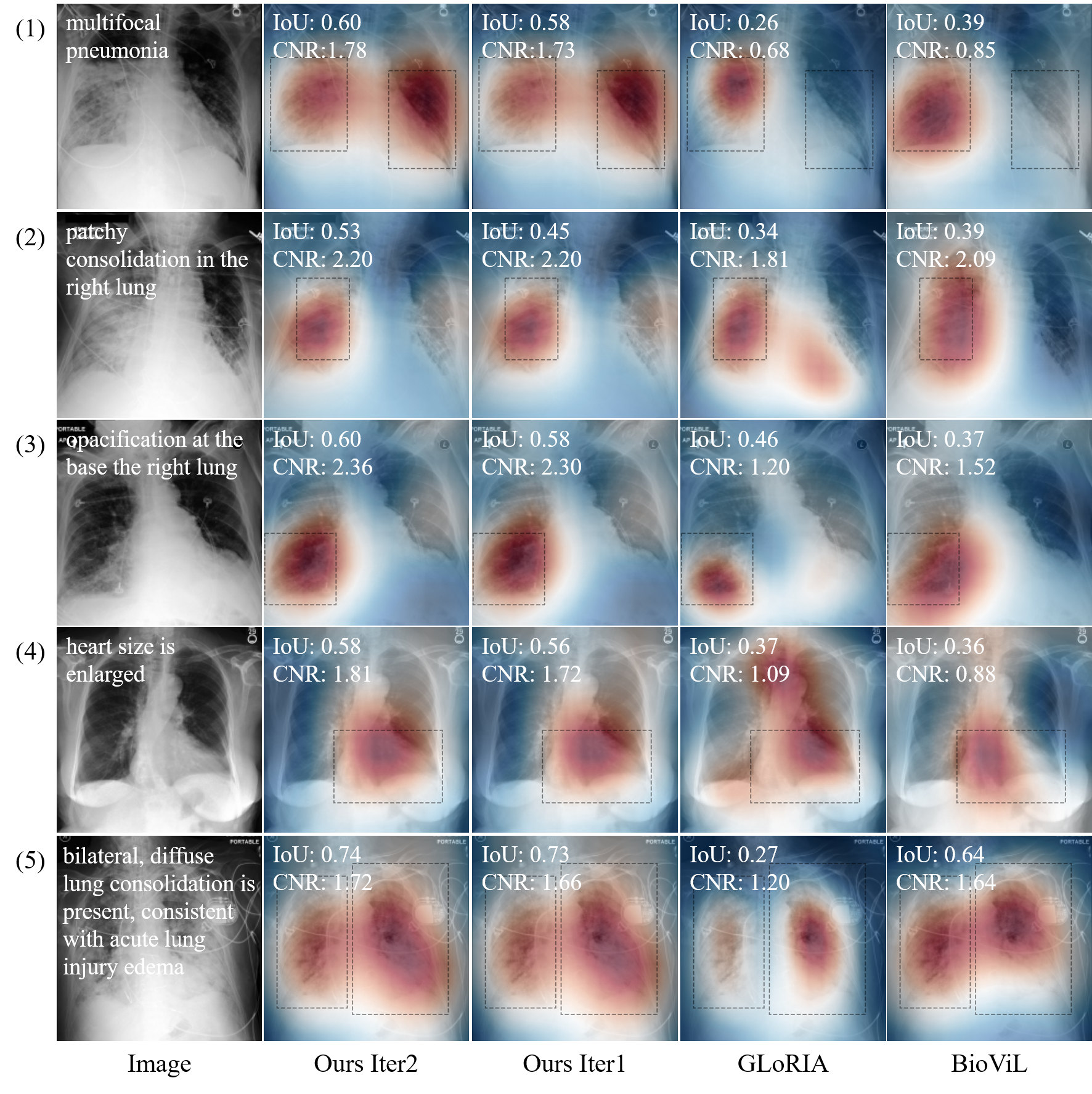}
\caption{Visualizations of phrase grounding with free text on the MS-CXR dataset. (1) to (5) represent five examples in the dataset. White color sentences are the provided free-text annotations. Dashed boxes indicate the annotations outlined by clinical experts. ``Ours Iter1'' and ``Ours Iter2'' represent the models trained after the first and second iterations in our framework, respectively.}
\label{fig3}
\end{figure*}

\subsection{Results for zero-shot phrase grounding}
\label{sec4.8}
Phrase grounding serves as a powerful tool for improving the interpretability of deep learning models. We evaluated the zero-shot phrase grounding performance of different methods using the MS-CXR dataset (Table \ref{tab5} and Fig. \ref{fig3}).

The quantitative results in Table \ref{tab5} demonstrate that our proposed framework can effectively enhance the zero-shot phrase grounding performance of the vision-language representation learning model, as evidenced by both evaluation metrics, CNR and mIoU. Particularly, the improvement in CNR is noteworthy. Specifically, our framework achieves a CNR improvement of 0.439, 0.327, and 0.230 when compared to ConVIRT, GLoRIA, and BioViL, respectively. CNR, which measures the difference between the similarity values of inner and outer bounding box regions without relying on hard threshold values, is a more objective metric that could be more clinically relevant, particularly when heatmap visualizations are needed instead of discrete segmentation \citep{Boecking2022BioVil}. In other words, our method can be very helpful in clinical applications where heatmap visualizations can provide more informative insights.

The example phrase grounding visualization results are presented in Fig. \ref{fig3}. For comparison, we included two well-known vision-language pretraining methods, GLoRIA and BioViL. Multifocal pneumonia is a common condition in clinical practice; however, as depicted in Fig. \ref{fig3}(1), GLoRIA and BioViL can not handle this scenario effectively. Additionally, GLoRIA often exhibits over-segmentation, as shown in Fig. \ref{fig3}(2) and Fig. \ref{fig3}(4). In contrast, both 'Ours Iter1' and 'Ours Iter2' demonstrate more stable performance in grounding tasks, with 'Iter2' slightly outperforming 'Iter1' in terms of IoU and CNR scores. These qualitative visualization results validate the effectiveness of our proposed framework in highlighting important regions for zero-shot tasks.

\subsection{Ablation study}
\label{sec4.9}
In this section, we present the results of our ablation studies, which aim to assess the effectiveness of each component in our method. To conduct a comprehensive evaluation, we performed both zero-shot and fine-tuning experiments. The zero-shot experiments include classification tasks on the NIH ChestX-ray dataset and the SIIM-ACR Pneumothorax dataset and phrase grounding tasks on the MS-CXR dataset. The fine-tuning experiments include classification tasks on the NIH ChestX-ray dataset and segmentation tasks on the SIIM-ACR pneumothorax dataset. During the fine-tuning process, we utilized All fine-tuning experiments were performed by utilizing 1\% labeled data. The results are listed in Table \ref{tab6}.

\begin{table*}[h]
\tiny
\begin{center}
\caption{Results of ablation studies. The influence of different module combinations and different dictionary construction methods is analyzed.}
\label{tab6}
\begin{tabular}{c|c|c|c|c|c}
\hline
\multirow{5}{*}{Methods} & \multicolumn{3}{|c|}{Zero-shot experiments} & \multicolumn{2}{|c}{Fine-tuning experiments} \\
% \cline{2-6}
~   & Classification & Classification & Classification & Classification & Segmentation\\
~   & (AUC: \%) & (AUC: \%)& (AUC: \%) & (AUC: \%) & (Dice: \%)\\
% \cline{2-6}
~   & NIH ChestX-ray & MS-CXR &SIIM-ACR  & NIH ChestX-ray & SIIM-ACR\\
\hline
Base model             & 72.6  & 65.9& 77.6  & 78.0 & 60.2 \\
Base+ITM               & 72.9  & 67.0& 79.7  & 77.9 & 60.7  \\
Ours Iter1             & 74.7  & 70.8& 87.8   & 78.5 & 61.2 \\
Ours Iter2 (GPT-4o)    & 75.7  & 72.2& 88.0  & 78.6 & \textbf{61.4} \\
Ours Iter2 (Claude-3)  & 75.5 & \textbf{73.1} & 88.5  & 78.6 & 61.3 \\
Ours Iter2 (Manual)    & \textbf{76.1}   & 73.0 & \textbf{88.5}& \textbf{78.7} &61.3\\
\hline
% \bottomrule
\end{tabular}
\end{center}
%\resizebox{\linewidth}{!}{
\end{table*}

\begin{figure*}[htb]
\centering
\includegraphics[scale=.35]{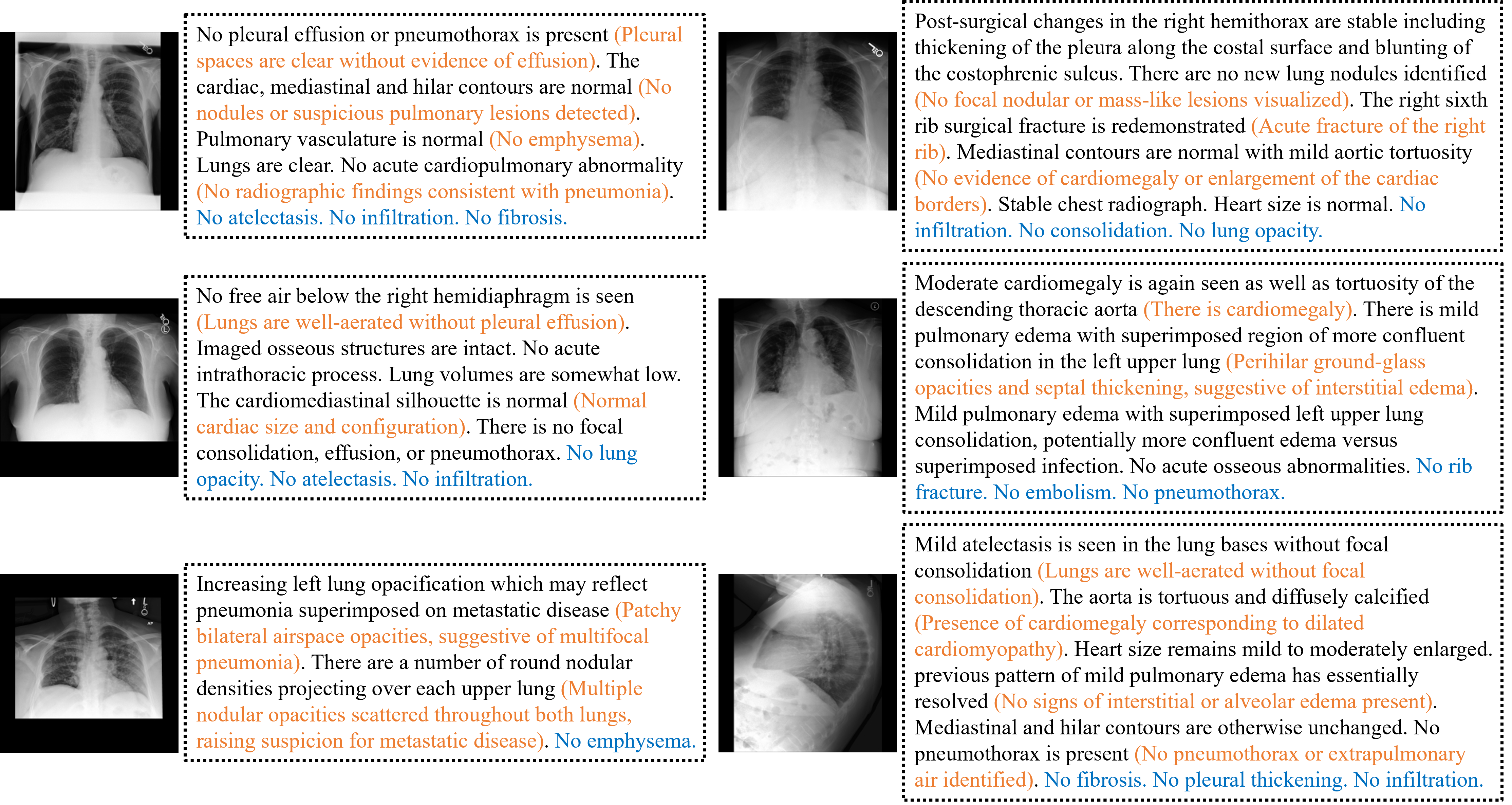}
\caption{Example refined reports of ``Ours Iter2 (Claude-3)''. The yellow sentences in parentheses represent the supplement sentences via the dictionary matching. The sentences in blue are negative sentences introduced to provide additional information by explicitly stating the absence of these diseases.}
\label{fig4}
\end{figure*}

We began our ablation studies with the base model, which utilized solely global contrastive learning without the symmetric local contrastive loss and ITM. Then, we added ITM (``Base+ITM'' in Table \ref{tab6}) to check the influence of image-text matching on the model performance. Next, we further incorporated the symmetric local contrastive loss (``Ours Iter1'' in Table \ref{tab6}), which formed our vision-language representation learning model with image-text matching determination capability (Fig. \ref{fig2}(a)). Afterward, we introduced the report refinement method and trained our model in two iterations. As demonstrated in Sec. \ref{sec3.2.2}, we incorporated several sentence generation methods, including: 1) using manually designed base templates (``Ours Iter2 (Manual)"), 2) generating sentences based on the large language model Claude-3-Sonnet with the highest similarity score (``Ours Iter2 (Claude-3)"), and 3) generating sentences based on the large language model GPT-4o with the highest similarity score ("Ours Iter2 (GPT-4o)").

The results presented in Table \ref{tab6} highlight the importance of each component in our method. ITM can improve the results of the zero-shot experiments, while its performance in the fine-tuning experiments is comparable to that of the base model. The introduction of the symmetric local contrastive loss in ``Ours Iter1'' leads to notable improvements across the majority of the tasks. Furthermore, our iterative learning model, combined with report refinement (Manual, Claude-3-Sonnet, GPT-4o), yields superior performance across all the experiments. In Fig. \ref{fig4}, we provide examples of refined reports generated by ``Ours Iter2 (Claude-3)''. Notably, the supplemented sentences accurately identify the key information from the original sentences while maintaining simplicity. This showcases the effectiveness of our approach in distilling complex information into concise and easily understandable sentences.

\begin{figure*}[htb]
\centering
\includegraphics[scale=.25]{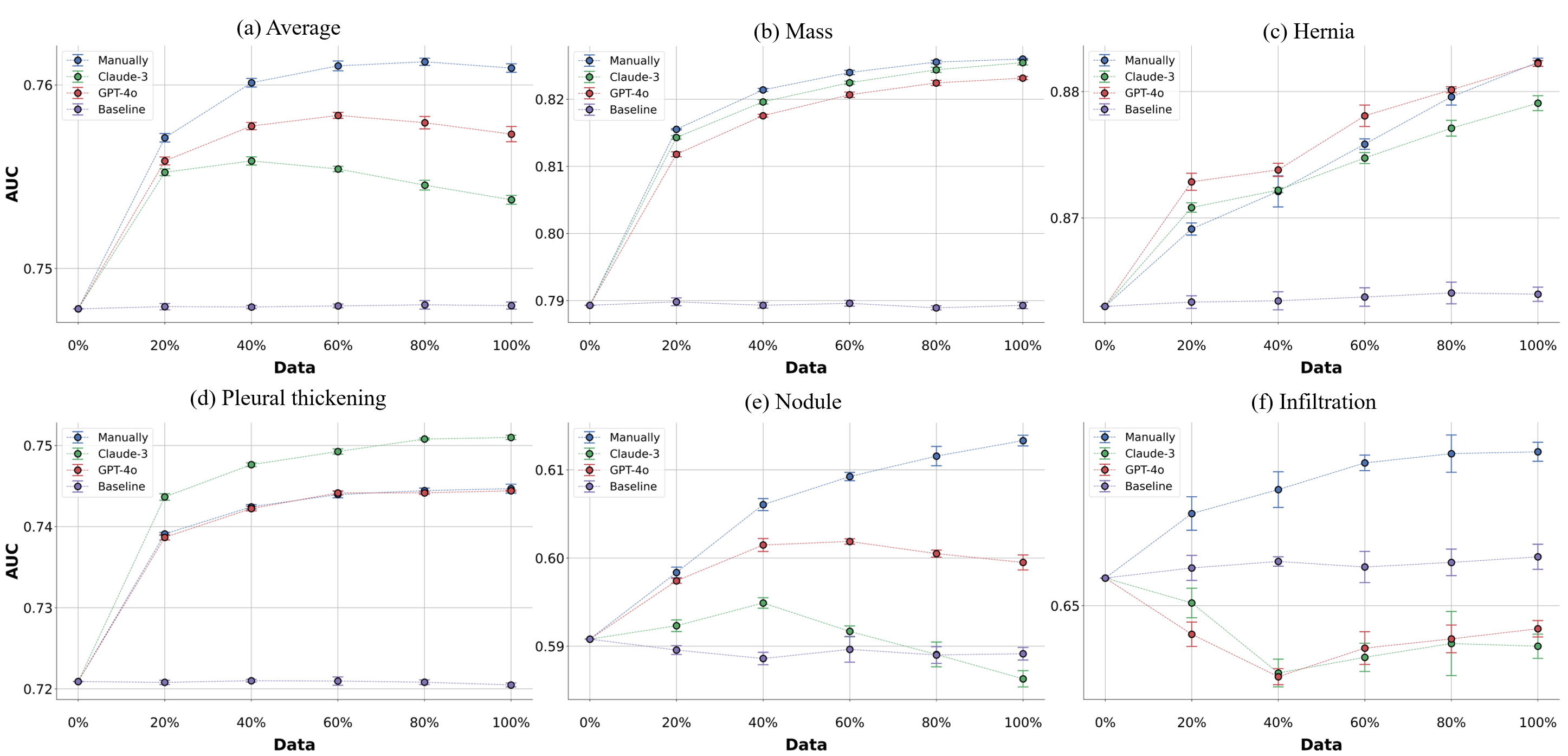}
\caption{Results of ablation studies by fine-tuning with different ratios of refined reports. The zero-shot classification results on the NIH dataset are reported, including the average scores for 14 diseases (a) and the specific scores for five diseases (b-f).}
\label{fig5}
\end{figure*}

To further analyze the effectiveness of the proposed report refinement and the differences among various dictionary construction methods, we fine-tuned the model using various refinement ratios in iteration 2. Specifically, 6 refinement ratios were experimented with, including 0\%, 20\%, 40\%, 60\%, 80\%, and 100\%. For fair comparison, we also report the fine-tuning results using different ratios of the original reports (denoted as ``Baseline''). Average zero-shot classification results on the NIH dataset as well as the results for five specific diseases are reported (Fig. \ref{fig5}). Following BioViL \citep{Boecking2022BioVil}, we performed five runs with different seeds. 

Several interesting findings were observed. For the average results (Fig. \ref{fig5}a), the baseline results showed no significant changes with increasing refinement ratios, indicating that the model of ``Iter1'' had already converged, and further training would not yield additional benefits. On the other hand, the proposed methods (Manually, Claude-3, GPT-4o) achieved further improved performance across different refinement ratios, underscoring the importance of report refinement. Besides, we found that the performance gains were not monotonically increasing. For example, with the manually constructed dictionary, the highest scores were achieved at the refinement ratio of 80\%, while with the dictionary created using Claude-3-Sonnet, the refinement ratio of 40\% became the best. One possible reason for this discrepancy could be that, although we generated direct disease-related sentences to enhance the model's capability using the language model, these sentences might also include imprecise matching ones, leading to degraded performance. This suggests that our dictionary construction and sentence selection methods still have room for further improvement. Similar phenomena were observed for specific diseases. Encouragingly, for mass, hernia, and pleural thickening, the performance improved continually with the increased adoption of refinement reports. Among the different dictionary construction methods, Claude-3-Sonnet particularly excelled in pleural thickening, likely because the dictionaries generated for this disease can better align with clinical practice, enhancing model learning (Appendix Table S1). Conversely, for nodule and infiltration, while the manually constructed dictionary maintained high performance, Claude-3-Sonnet (for nodule) and GPT-4o (for both nodule and infiltration) exhibited performance declines. This could be caused by the inability of these large language models to comprehensively capture the clinical scenarios for these diseases (Appendix Table S1 and Table S2). These experiments indicate that the proposed report refinement method can effectively enhance the medical vision-language model's ability to understand chest X-ray diseases. Although current language models can not yet outperform manually constructed dictionaries in overall performance, they have shown comparable or superior performance for certain specific diseases.

The above experiments indicate that the proposed report refinement method can effectively enhance the medical vision-language model's ability to understand chest X-ray diseases. Although the sentences generated by large language models (Claude-3-Sonnet and GPT-4o) provide more detailed descriptions, they may not be precisely aligned with the images and could introduce incorrect information. In contrast, manually generated sentences focus on conveying the presence or absence of diseases, sacrificing some detail but ensuring the accuracy of the supplemented information. In this study, we retained the original sentences in the report during the refinement process. By combining the original sentences with the manually generated ones, we achieve a balance between providing detailed descriptions and offering straightforward guidance regarding the presence of diseases. This approach ensures both comprehensive information and clear indications of disease presence, thereby enhancing the overall understanding of the radiology reports. The method labeled as ``Ours Iter2 (Manual)'' with 100\% report refinement ratio is our final approach. All results reported in Tables \ref{tab1} to \ref{tab5}, labeled as ``Ours'', are achieved using this method.

\section{Discussion and Conclusion}
In this study, we developed a novel iterative vision-language representation framework that incorporates a key semantic knowledge-emphasized report refinement method. The primary objective of our work was to enhance the representation learning process by refining complex radiology reports. Extensive experiments were conducted on five external datasets, encompassing different medical image recognition tasks and evaluation settings, to investigate the effectiveness of the proposed framework. The results consistently demonstrated that our framework outperformed existing state-of-the-art methods, showcasing its superiority and robustness.

Recently, the rapid development of large language models has brought new opportunities to the medical field. We also compared medical dictionaries constructed based on the latest advanced language models. Although these language models demonstrated suboptimal performance in our tests, we believe that with the advancement of multimodal medical models, more comprehensive and fine-grained disease descriptions from language models will emerge in the future, further enhancing the performance of our model.

%potential limitations
Despite obtaining promising results, our framework still has the following limitations. Firstly, the performance of the report refinement method relies on the constructed clinical dictionary, which can eventually impact the performance of the optimized model in downstream tasks. Currently, the dictionary was constructed semi-automatically, involving traversing all the training data. Fortunately, as shown in the supplementary file (Appendix Table S1 and Table S2), the dictionary is relatively simple and straightforward to construct. Secondly, as the two model-optimized knowledge-enhancement metrics (similarity metric and matching metric) rely on Stage 1 trained image and text encoders, our framework needs to be trained in two iterations. In future research, we will explore methods for integrating these different steps and training the framework end-to-end.

In summary, our proposed key semantic knowledge-emphasized report refinement method effectively refined the complex radiology reports to highlight crucial information. Leveraging these refined reports, our iterative vision-language representation learning framework enables effective utilization of knowledge within radiology reports and facilitates the learning of meaningful medical image representations for various downstream medical image analysis tasks, including fine-tuning disease classification, fine-tuning region-of-interest segmentation, zero-shot disease classification, and zero-shot phrase grounding. The consistent improvement in performance across different tasks highlights the potential of our framework as a valuable medical foundation model for diverse clinical applications.

\section*{Acknowledgments}
This research was partly supported by the National Natural Science Foundation of China (No. 62222118, No. U22A2040),
Guangdong Provincial Key Laboratory of Artificial Intelligence in Medical Image Analysis and Application (No. 2022B1212010011), 
Shenzhen Science and Technology Program (No. RCYX20210706092104034, No. JCYJ20220531100213029),
and Youth lnnovation Promotion Association CAS.

%%Harvard
% \bibliography{Vision_language_foundation_model}% common bib file
%%Harvard
\bibliography{Vision_language_foundation_model}

\end{document}